\documentclass[twocolumn]{article}

\usepackage{graphicx} 
\usepackage{amsmath} 
\usepackage{amssymb} 
\usepackage{cite} 
\usepackage{subcaption} 
\usepackage{float} 
\usepackage{hyperref} 
\usepackage[numbers]{natbib} 

\title{5G LDPC Linear Transformer for Channel Decoding}

\author{
    \begin{tabular}{cc}
        \textbf{Mario Hernandez} & \textbf{Fernando Pinero} \\
        University of Puerto Rico & University of Puerto Rico \\
        \texttt{mario.hernandez4@upr.edu} & \texttt{fernando.pinero1@upr.edu} \\
    \end{tabular}
}

\date{} 

\begin{document}

\maketitle

\begin{abstract}
    \textbf{This work introduces a novel, fully differentiable linear-time complexity transformer decoder and a transformer decoder to correct 5G New Radio (NR) LDPC codes\footnote{All code is available at: \href{https://github.com/pollyjuice74/5G-Decoder}{\texttt{https://github.com/pollyjuice74/5G-Decoder}}.}. We propose a scalable approach to decode linear block codes with \(O(n)\) complexity rather than \(O(n^2)\) for regular transformers. The architectures' performances are compared to Belief Propagation (BP), the production-level decoding algorithm used for 5G New Radio (NR) LDPC codes. We achieve bit error rate performance that matches a regular Transformer decoder and surpases one iteration BP, also achieving competitive time performance against BP, even for larger block codes. We utilize Sionna\cite{sionna2022}, Nvidia's 5G \& 6G physical layer research software, for reproducible results.}
\end{abstract}

\section{Introduction}

In the age of information, robust digital communication has become a cornerstone of modern wireless networks, ensuring data integrity across noisy transmission channels. Central to achieving reliability is the design of efficient error-correcting decoders that balance decoding time and error-correcting capabilities to provide practical solutions as we approach maximum likelihood (ML) decoding \cite{fossorier1995soft}, a theoretical optimum constrained by NP-hard complexity. Belief propagation decoding has emerged as the standard approach for 5G New Radio LDPC codes due to its iterative message-passing scheme and low computational overhead, offering a flexible number of iterations for varying channel conditions and data integrity requirements. While practical, it struggles to match the performance of ML decoding, particularly for short and intermediate block lengths—an ongoing challenge in modern communication systems.

Recent breakthroughs in deep learning have introduced new possibilities for error correction and communication \cite{wiesmayr2024design, dorner2017deep, gruber2017deep, nachmani2016learning, oshea2017introduction}. Among them, graph neural networks \cite{cammerrer2022gnnchannel, agrawal2017machine} have been proposed for their scalability and low-overhead integration into belief propagation decoding, leveraging parameterized iterative decoding. Error-correcting code transformers \cite{choukroun2022error} represent a notable advancement in decoding architectures, designed as unified solutions for linear block codes. These models leverage the attention mechanism, which operates effectively as a fully connected graph, capturing intricate relationships between codeword elements. This graph-based structure aligns naturally with the decoding requirements of linear block codes, which are traditionally decoded using graph-structured mechanisms like Tanner graphs.

However, transformers are limited in scalability due to their exponential training complexity as code size increases. Consequently, their practical application is constrained, particularly for the larger block sizes required in digital communication.

To address these challenges, we propose a novel transformer-based error correction decoder designed to overcome the scalability constraints of existing deep learning decoders. Our approach integrates domain knowledge by incorporating the code’s parity-check matrix into the self-attention mechanism, as described in \cite{choukroun2022error}, effectively guiding the learning process while maintaining computational efficiency. By utilizing Nvidia’s Sionna 5G \& 6G Physical Layer Research Software \cite{sionna2022}, we ensure reproducible results across varying parity-check matrix sizes, enabling a fair evaluation of our method’s performance.

A key innovation in our work is the introduction of a linear transformer decoder, using the attention proposed in \cite{wang2020linformer}, that significantly reduces the computational complexity from quadratic $O(n^2)$ to linear $O(n)$ scaling. This improvement enables our approach to efficiently decode large block sizes without an exponential increase in training and inference time. To the best of our knowledge, this is the first transformer-based decoder that is not constrained by the curse of dimensionality when applied to 5G NR LDPC codes and other linear block codes. Our contributions can be summarized as follows:

\begin{itemize}
    \item We introduce transformer decoders for 5G NR LDPC codes.
    \item We introduce a fully differentiable linear complexity transformer-based decoder for linear binary block codes.
    \item We provide a benchmark comparison against state-of-the-art 5G NR LDPC decoders.
    \item We present a hyper-parameter optimization study to guide future research on transformer-based decoding.
    \item We ensure reproducible research by implementing our solution within the Sionna framework and making our code publicly available.
\end{itemize}

This work aims to bridge the gap between deep learning and structured decoding algorithms, paving the way for future advancements in AI-driven error correction for next-generation communication systems.

\section{Communication Channels}

\subsection{Encoding and Modulation}
The encoding process maps information bits $b$ to a codeword $c$ using an encoder $\mathcal{E}$:
\begin{equation}
    c = \mathcal{E}(b)
\end{equation}
For uncoded transmission, $c = b$. The encoded bits are then mapped to complex symbols using a QAM mapper $\mathcal{M}$:
\begin{equation}
    x = \mathcal{M}(c)
\end{equation}
If $n$ is odd, zero-padding is applied to $c$ before mapping.

\subsection{Bipartite Graph Structure of the Parity Check Matrix}
Linear block codes such as Low-Density Parity-Check (LDPC) codes are often represented using bipartite graphs that consists of two sets of nodes: variable nodes (VNs) and check nodes (CNs). The edges connect VNs to CNs based on the parity-check matrix $H$. For a codeword $c$ and parity-check matrix $H$, the relation is given by:
\begin{equation}
    Hc^T = 0
\end{equation}
where $H$ is an $(n-k) \times n$ matrix over $\mathbb{GF}(2)$, representing the parity-check constraints, thus a recieved codeword can be decoded.

\subsection{Channel Model}
The modulated symbols \( x \) are transmitted through an Additive White Gaussian Noise (AWGN) channel:
\begin{equation}
    y = x + w, \quad w \sim \mathcal{N}(0, N_0)
\end{equation}
where the noise variance \( N_0 \) is computed from the given \( E_b/N_0 \) using the method described in \cite{sionna2022ebnodb2no}:
\begin{equation}
    N_0 = \left(\frac{E_b}{N_0} \cdot \frac{rM}{E_s}\right)^{-1}
\end{equation}
For coded transmission, the term \( M \) represents the average number of coded bits per constellation symbol, \( E_s \) is the average energy per symbol, and \( r \) is the coderate. For uncoded transmission, the coderate \( r \) is set to 1.

\subsection{Demodulation and Decoding}
The received signal \(y\) is converted into LLRs using a demapper \(\mathcal{D}\):
\begin{equation}
    \Lambda(c) = \mathcal{D}(y, N_0) \in \mathbb{R}^{n \times 1}
\end{equation}
Here, \(\Lambda(c)\) represents the log-likelihood ratio (LLR), quantifying the confidence in each bit \(c\) being \(0\) or \(1\), bounded by \(\pm 20\).

For syndrome-based decoding, the decoder input nodes are:
\begin{equation}
    \sigma = H c^T \in \{0,1\}^{m \times 1},
\end{equation}
where $H$ is the parity-check matrix. The decoder input is then formed as the following conatenation:
\begin{equation}
    X_{\text{nodes}} = \begin{bmatrix} \Lambda(c) \\ \sigma \end{bmatrix} \in \mathbb{R}^{(n+m) \times 1}
\end{equation}

The decoder $\mathcal{D}_e$ estimates the transmitted codeword:
\begin{equation}
    \hat{c} = \mathcal{D}_e(X_{\text{nodes}})
\end{equation}
where the estimated bits are obtained by thresholding:
\begin{equation}
    \hat{c}_i = \begin{cases} 1, & \text{if } \hat{c}_i > 0 \text{ (logit threshold)} \\ 0, & \text{otherwise} \end{cases}
\end{equation}

\section{LDPC Codes}

Here we illustrate the parity check matrix and mask, as proposed in \cite{choukroun2022error}, for regular LDPC and 5G New Radio LDPC. We also point out the difference in construction.

\begin{figure}[H]
    \centering
    \begin{subfigure}[t]{0.8\linewidth}
        \centering
        \includegraphics[width=\linewidth]{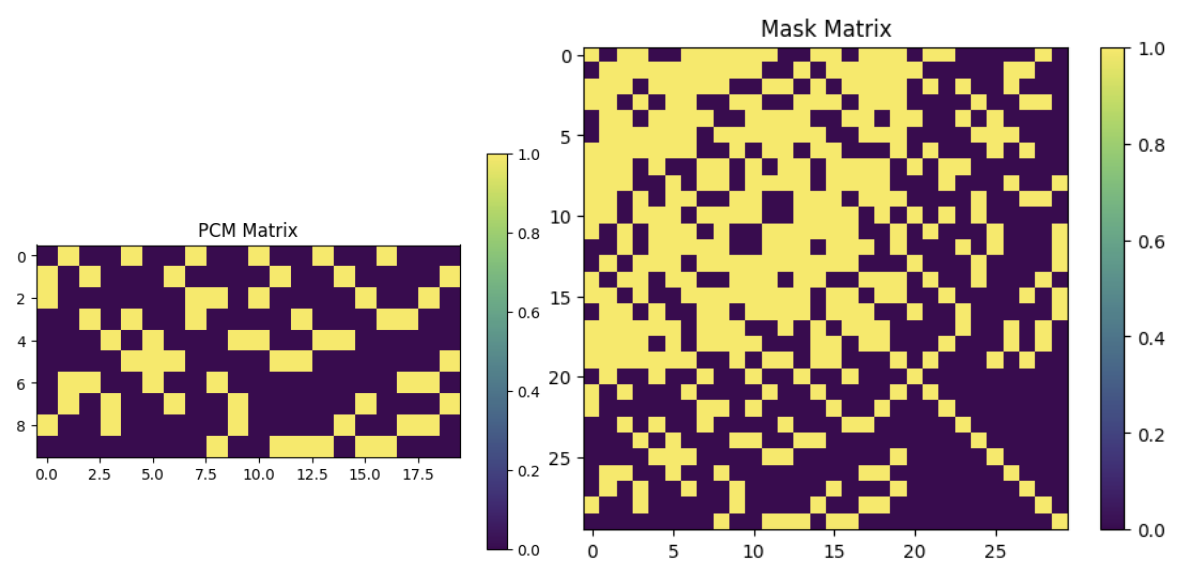}
        \caption{Regular LDPC \textbf{parity check matrix} and \textbf{mask}. PCM with linearly independent rows, each row has c=6 check node degrees and each column has v=3 variable node degrees.\cite{ryan2004ldpc, gallager1963ldpc}}
        \label{fig:reg-LDPC}
    \end{subfigure}%
    \hspace{0.02\linewidth} 
    \begin{subfigure}[t]{0.8\linewidth}
        \centering
        \includegraphics[width=\linewidth]{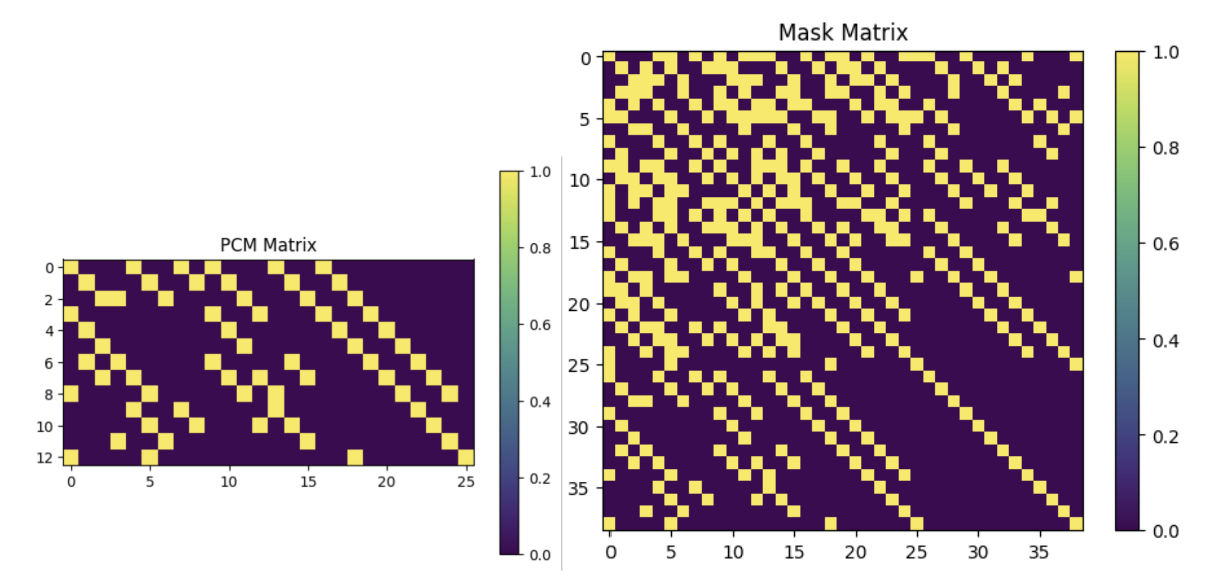}
        \caption{5G New Radio LDPC \textbf{parity check matrix} and \textbf{mask}. PCM constructed by lifting the selected base graph. \cite{wiesmayr2024puncturing}}
        \label{fig:5G-LDPC}
    \end{subfigure}
    \label{fig:LDPC-comparison}
\end{figure}

\subsection{Regular LDPC Code Construction}
A regular \((v, c)\) Low-Density Parity-Check (LDPC) code is constructed by defining three key parameters \cite{sionna_ldpc}: \(v\), the number of edges connected to each Variable Node (VN); \(c\), the number of edges connected to each Check Node (CN); and \(n\), the codeword length. These parameters determine the structure and connectivity of the LDPC code, ensuring that it adheres to the regularity constraints specified by the \(v\) and \(c\) values.

The number of check nodes is determined as:
\begin{equation}
    n_c = \frac{v}{c} n,
\end{equation}
where the code rate is:
\begin{equation}
    R = 1 - \frac{v}{c}.
\end{equation}
The parity-check matrix $H$ is formed by randomly assigning connections between VNs and CNs while maintaining the defined node degrees. The construction does not optimize for cycle avoidance, and permutations are applied to improve randomness. The resulting $H$ matrix is then used for encoding and decoding processes.

\subsection{5G LDPC Rate-Matching}
5G NR codes utilize rate matching to fit the length of the codeword to the desired transmission length. Rate matching is performed using three techniques: shortening, pruning, and puncturing \cite{richardson2018ldpc, hui2018channelcoding}. In a scenario without rate matching, the code operates as a regular LDPC decoder with a parity-check matrix (PCM) and generator matrix derived from the base graph specified for the chosen code rate and codeword length.

\subsection{Puncturing}
A $(k,n)$ mother code is punctured by not transmitting certain codeword bits. This increases the code rate \cite{sionna5gchannelcoding}:
\begin{equation}
    r_{\text{pun}} = \frac{k}{n - p} > \frac{k}{n}, \quad \forall p > 0.
\end{equation}
At the decoder, these punctured codeword bits are treated as erasures ( $\ell_{\text{ch}} = 0$).

\subsection{Shortening}
A $(k,n)$ mother code is shortened by setting $s$ information bits to fixed (known) values. Assuming systematic encoding, these $s$ positions are not transmitted, leading to a new code rate:
\begin{equation}
    r_{\text{short}} = \frac{k - s}{n - s} < \frac{k}{n}.
\end{equation}
At the decoder, these shortened codeword bits are treated as known values ($\ell_{\text{ch}} = \infty$).

By removing rate matching, the encoder directly maps information bits into codewords using the generator matrix. The decoder utilizes the PCM to correct errors by ensuring the syndrome remains zero. This approach simplifies LDPC decoding while maintaining high reliability and flexibility in 5G NR systems.

\section{Transformers for channel decoding}

The Transformer model, introduced by \cite{vaswani2017attention}, is a neural network architecture that relies entirely on self-attention mechanisms to capture global dependencies between input and output. 

Our model first embeds the input, resulting in a shape of \((\text{batch size}, n+m, \text{hidden dims})\), and then passes it through \(N\) transformer blocks, as illustrated below. Finally, it goes through a feed-forward and fully-connected layer, changing its shape from \((\text{batch size}, n+m)\) to \((\text{batch size}, n)\), respectively.

\begin{figure}[H]
    \centering
    \includegraphics[width=0.4\linewidth]{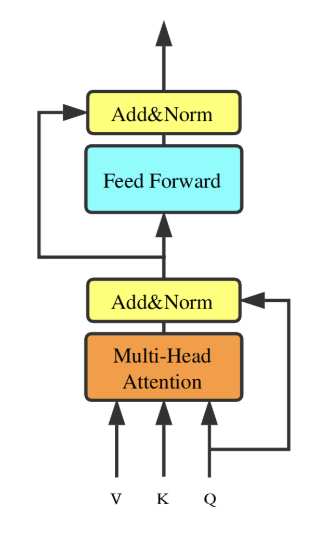}
    \caption{Transformer Block architecture}
    \label{fig:enter-label}
\end{figure}

\subsection{Attention}
The self-attention mechanism is defined as:
\begin{equation}
    \text{Attention}(Q, K, V) = \text{softmax}\left(\frac{QK^T}{\sqrt{d_k}}\right)V
\end{equation}
where $Q$, $K$, and $V$ are the query, key, and value matrices, respectively, and $d_k$ is the dimensionality of the keys.

\subsection{Linear Attention}
Linear attention is achieved by reducing the complexity of traditional softmax-based self-attention from \(O(n^2)\) to \(O(n)\). Our implementation is a version of the key-value linear attention introduced in \cite{wang2020linformer}. This is accomplished by projecting the key and value tensors into a low-dimensional space and applying key-value compression. Below, we outline the process with the associated tensor shapes at each step.

1. \textbf{Input Shapes}: The query, key, and value tensors are represented as:
   \[
   \mathbf{Q}, \mathbf{K}, \mathbf{V} \in \mathbb{R}^{B \times N \times D},
   \]
   where \(B\) is the batch size, \(N\) the sequence length, and \(D\) the model dimension.

2. \textbf{Projection into Low-Rank Space}: The key and value tensors are projected into a lower-dimensional space:
   \[
   \mathbf{K}' = \mathbf{K} \mathbf{P}_K, \quad \mathbf{V}' = \mathbf{V} \mathbf{P}_V,
   \]
   with \(\mathbf{P}_K, \mathbf{P}_V \in \mathbb{R}^{N \times K}\). Resulting shapes:
   \[
   \mathbf{K}' \in \mathbb{R}^{B \times H \times K \times D_H}, \quad \mathbf{V}' \in \mathbb{R}^{B \times H \times K \times D_H}.
   \]

3. \textbf{Scaled Dot-Product Attention}: Attention scores are computed as:
   \[
   \text{Scores} = \frac{\mathbf{Q} \cdot \mathbf{K}'^\top}{\sqrt{D_H}}, \quad \text{Scores} \in \mathbb{R}^{B \times H \times N \times K}.
   \]
   A low-rank mask of shape \((N, K)\) is resized and applied:
   \[
   \text{Scores}_{\text{masked}} = \text{Scores} + \text{Mask}.
   \]

The core idea behind linear attention can be observed here. In standard attention, without using projection matrices, the resulting score tensor has the shape \(\text{Scores} \in \mathbb{R}^{B \times H \times N \times N}\), leading to a time complexity of \(O(N^2)\). However, by introducing a feature dimension \(K\) such that \(K \ll N\), or by fixing \(K\) as a constant while allowing \(N\) to increase, the complexity reduces to \(O(NK)\). In the case where \(K\) is a fixed constant, this further simplifies to \(O(N)\), resulting in a linear complexity with respect to sequence length.

4. \textbf{Softmax and Weighted Summation}: Softmax is applied to normalize the scores:
   \[
   \text{A} = \text{softmax}(\text{Scores}_{\text{masked}}), \quad \text{A} \in \mathbb{R}^{B \times H \times N \times K}.
   \]
   The weighted summation with \(\mathbf{V}'\) produces:
   \[
   \text{Output} = \text{A} \cdot \mathbf{V}', \quad \text{Output} \in \mathbb{R}^{B \times H \times N \times D_H}.
   \]

5. \textbf{Final Reshaping and Output}: The output is reshaped to:
   \[
   \text{Output} \in \mathbb{R}^{B \times N \times D}.
   \]

\section{Results}

Our results are compared against one-iteration Belief Propagation decoding, as iterative transformer decoding has been proposed in \cite{choukroun2023denoising} through a diffusion-based method. Belief propagation is the primary decoding scheme for 5G New Radio (NR) LDPC codes due to its flexibility in adjusting decoding iterations for varying channel conditions and requirements. It is important to note that 5G NR LDPC codes are highly optimized for BP decoding. We also highlight that this is not an exact implementation of the transformer model proposed by \cite{choukroun2022error}, which naturally leaves room for potential architectural enhancements. Our decoders perform best for `small' PCM sizes (\(n = 20\text{--}150, r = 0.5\)) per 5G NR standards.

\begin{figure}[H]
    \centering
    \includegraphics[width=0.9\linewidth]{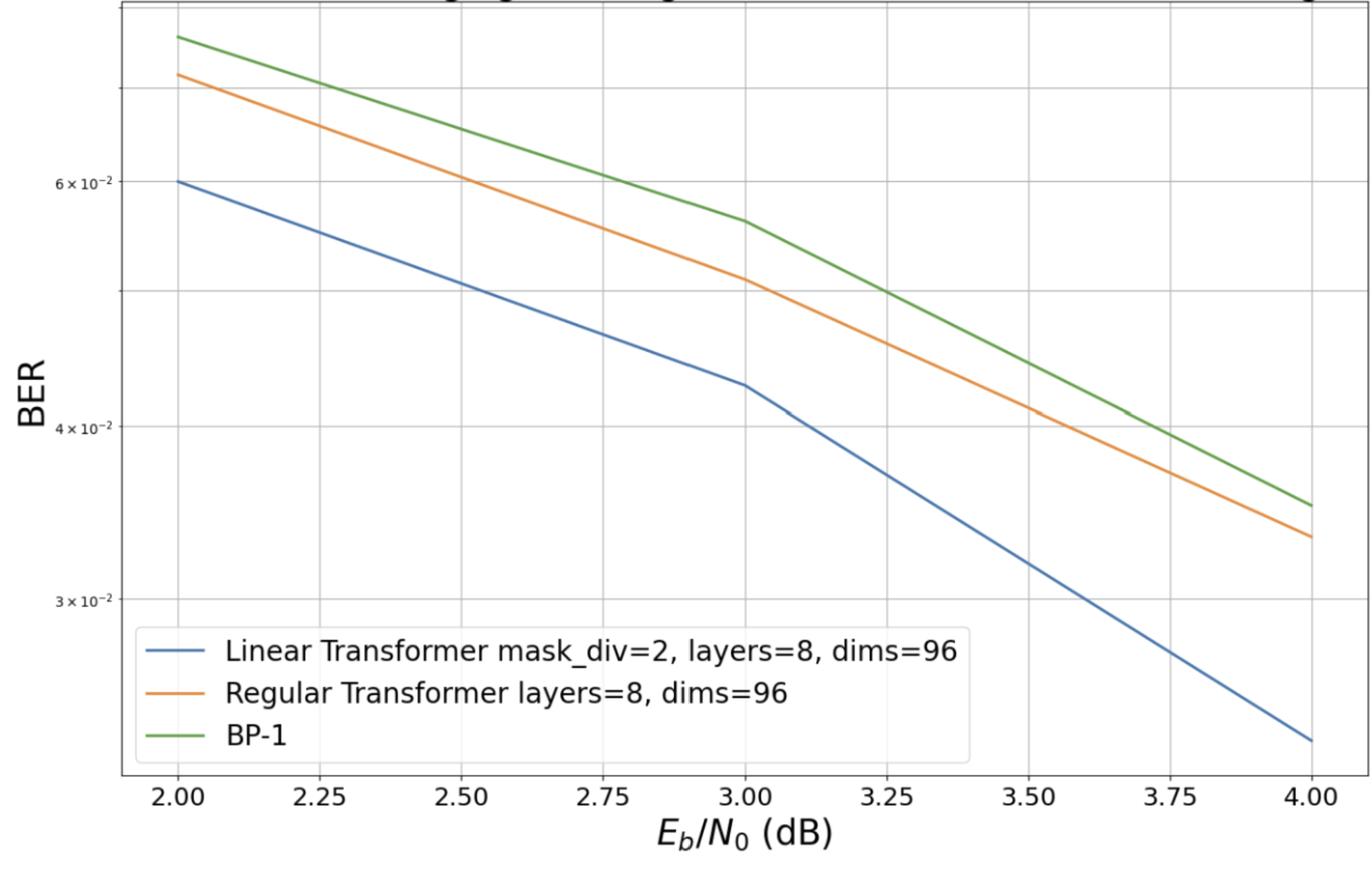}
    \caption{BER comparison for \((k, n) = (13, 26)\).}
    \label{fig:comparison-13-26}
\end{figure}

\begin{figure}[H]
    \centering
    \includegraphics[width=0.9\linewidth]{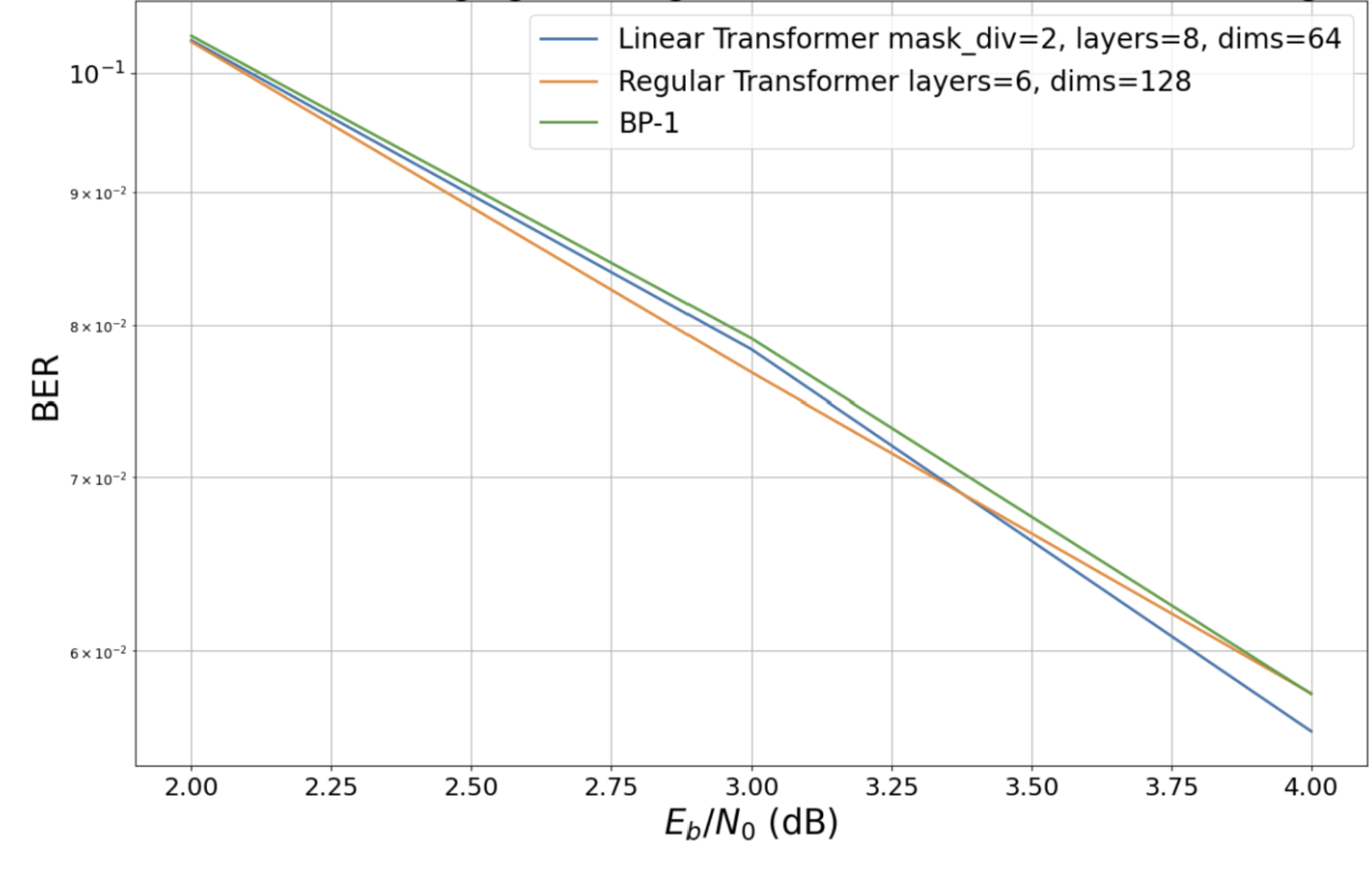}
    \caption{BER comparison for \((k, n) = (35, 82)\).}
    \label{fig:enter-label}
\end{figure}

Transformer and linear transformer-based decoding achieves superior performance than one-iteration belief propagation (BP) for 5G NR LDPC codes. Here, we use a mask division value of 2 for all linear transformer decoders. Both transformer and linear transformer models were trained for approximately 6 hours for each PCM shape. Linear transformers in these block sizes trained approximately three times faster, consequently allowing training for three times more iterations than regular transformers, thus achieving superior decoding performance. It is also worth noting that larger PCM shapes could not be tested due to limited GPU memory.

\begin{figure}[H]
    \centering
    \includegraphics[width=1\linewidth]{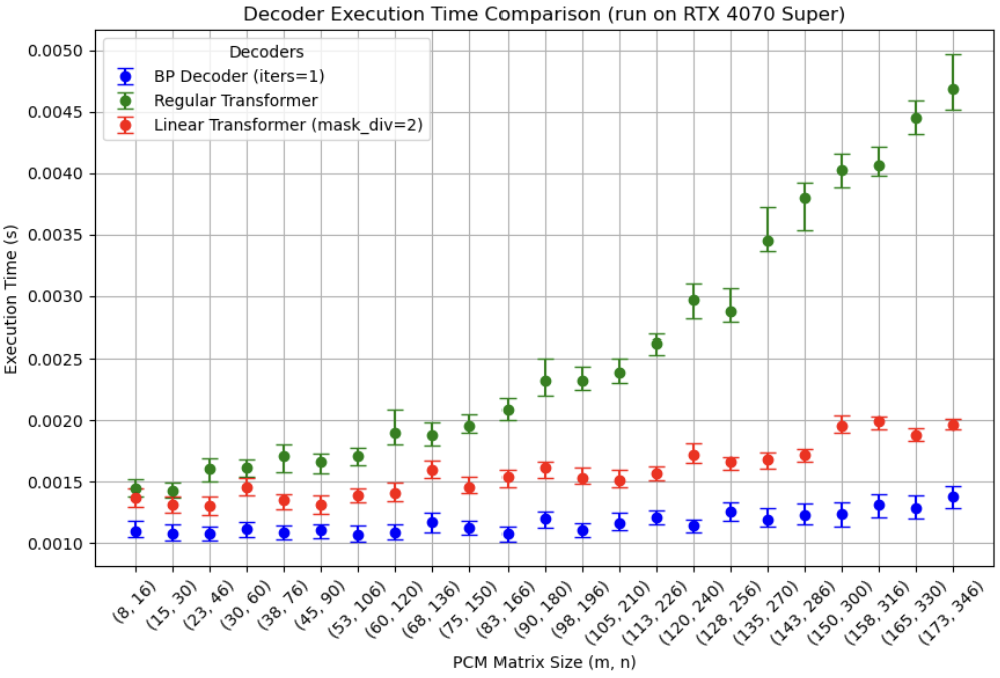}
    \caption{Time comparison between transformer, linear transformer and belief propagation decoding with differing block sizes of code rate r=0.5.}
    \label{fig:enter-label}
\end{figure}

These results demonstrate significant improvements in time complexity for linear transformer-based decoding compared to regular transformer-based decoding. GPU parallel processing enables horizontal scaling, meaning the time required for additional computations does not increase linearly with the number of iterations or input size. GPUs efficiently distribute workloads across thousands of cores, enhancing computational efficiency and making iterative methods a promising approach for future research.

\begin{figure}[H]
    \centering
    \includegraphics[width=1\linewidth]{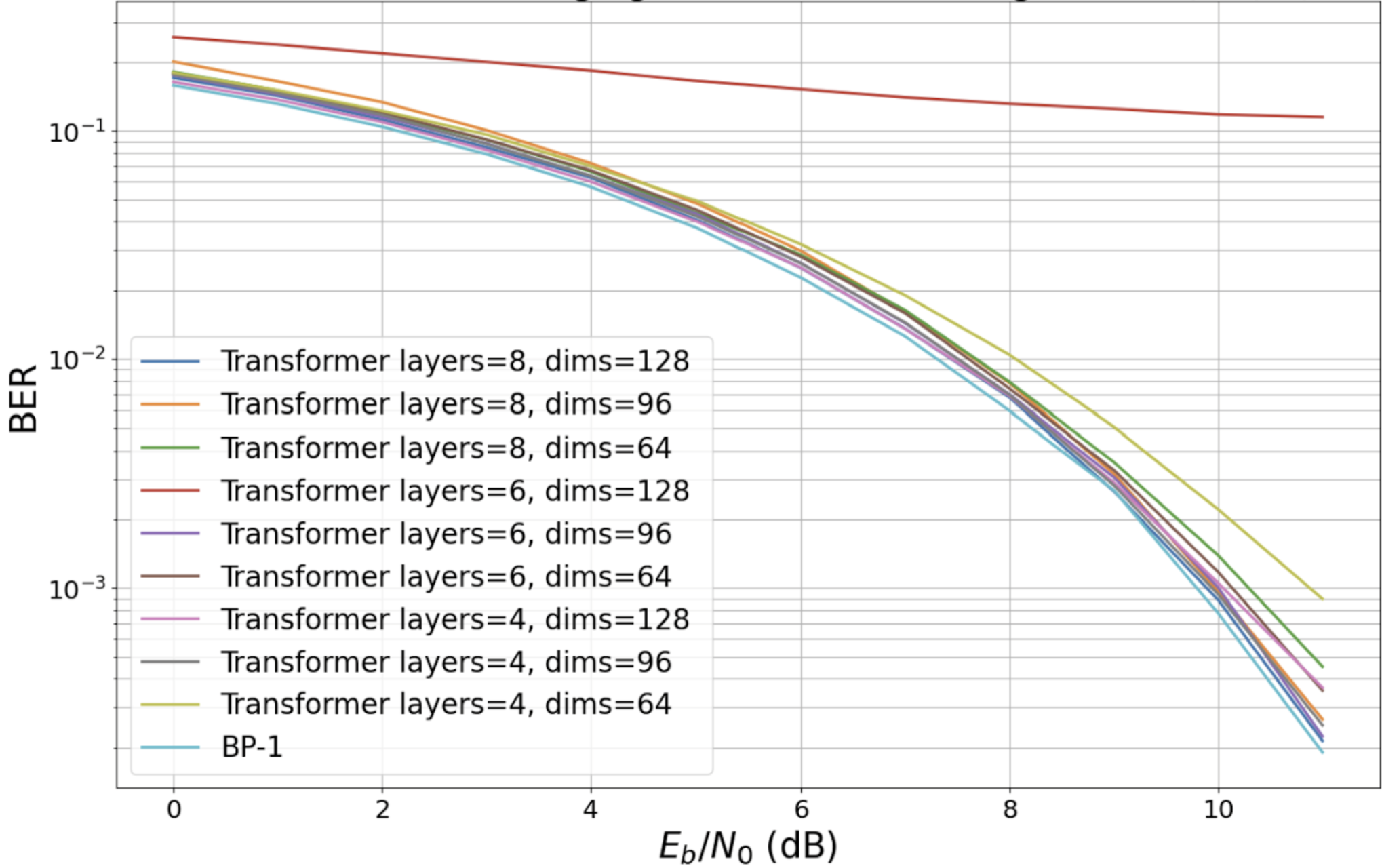}
    \caption{ \( (k, n) = (256, 576)\).}
    \label{fig:comparison-256-576}
\end{figure}

\begin{figure}[H]
    \centering
    \includegraphics[width=1\linewidth]{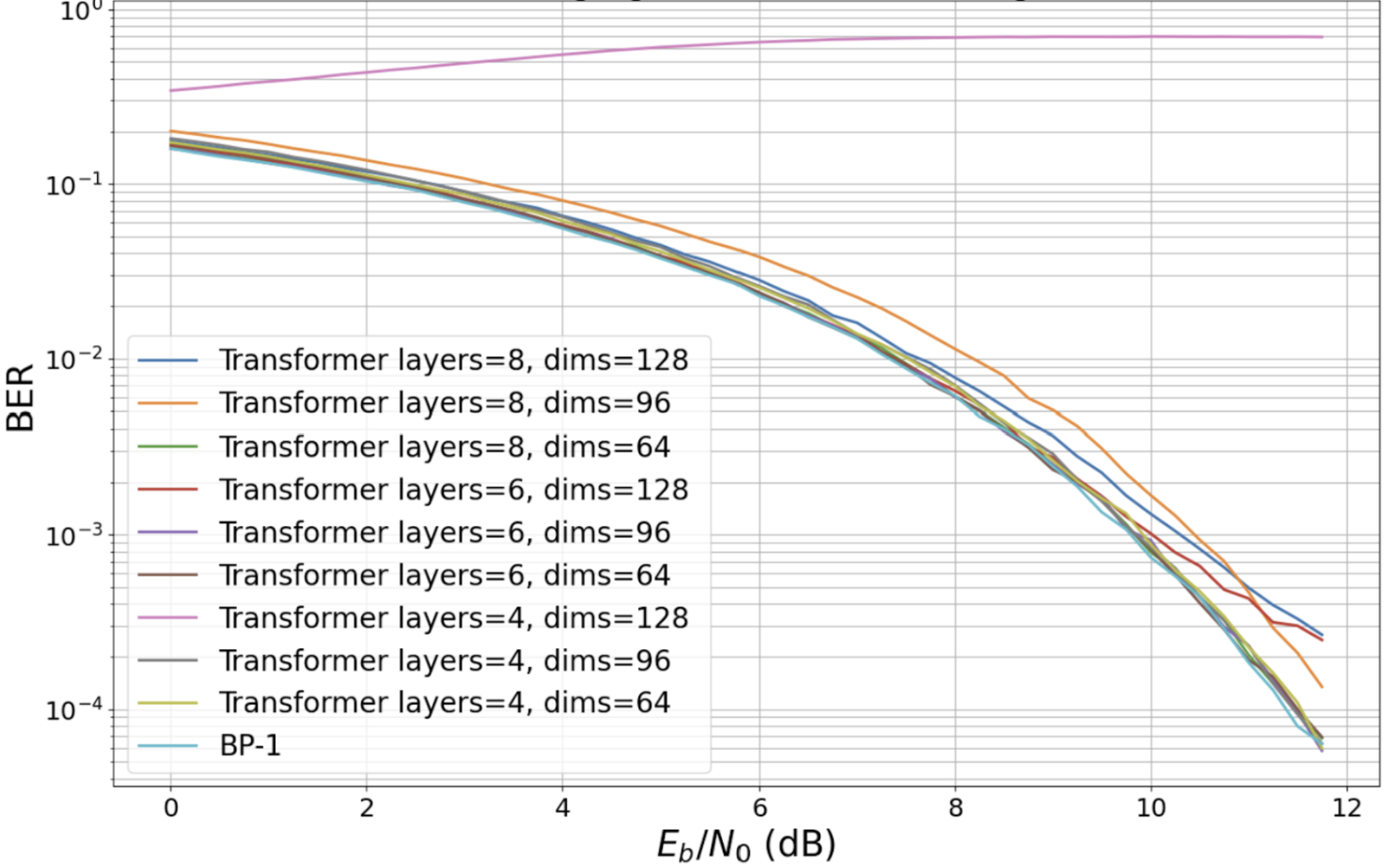}
    \caption{\((k, n) = (192, 448)\).}
    \label{fig:comparison-192-448}
\end{figure}

\begin{figure}[H]
    \centering
    \includegraphics[width=1\linewidth]{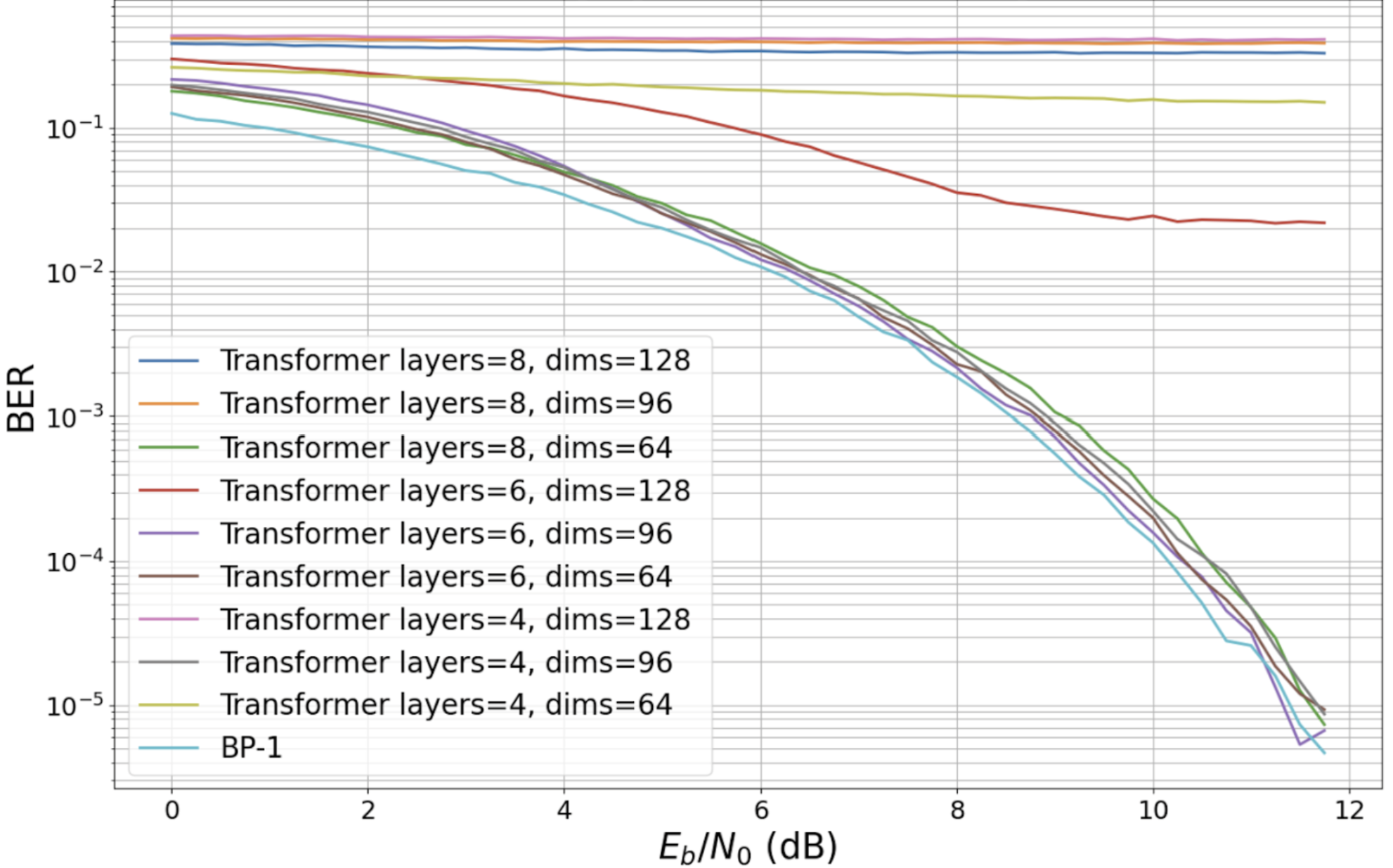}
    \caption{\((k, n) = (100, 186)\).}
    \label{fig:comparison-100-186}
\end{figure}

Here we compare different hyper-parameters for various PCM shapes \((k, n)\) during training. All models were trained for 1000 iterations (around 8 minutes per model on our GPU) with \(E_b/N_0\) values between 8--15 dB and a learning rate of \(5 \times 10^{-3}\). 

Transformer models converge very quickly to one-iteration BP. Further fine-tuning requires learning rate adjustments; we used cosine learning rate decay. It is worth noting that the model appears to generalize well for correcting \(E_b/N_0\) values, achieving faster and better results when trained on higher dB values. We suggest initially training in a larger dB range (0--8) and subsequently fine-tuning on higher dB values (4--8).

\section{Conclusion}
We present a novel Linear Transformer and Transformer architecture for decoding 5G NR LDPC codes and regular LDPC codes, addressing traditional scalability and efficiency challenges with Transformer models. Using linear attention, we successfully reduce the complexity from \(O(n^2)\) to \(O(n)\), allowing the decoding of larger block sizes with a competitive bit error rate. Our results demonstrate performance comparable to one-iteration belief propagation, the current production standard for 5G NR LDPC decoding. We have shown that Transformer models converge extremely quickly (within 1000 training iterations) to one-iteration BP. Our work opens the door for future advancements, including iterative transformer decoding for 5G codes and transformer architectural enhancements.

\section{Acknowledgements}
We thank NASA Space Grant and the National Science Foundation for funding our work. Boqueron HPCf for their scientific computation cluster.

\bibliographystyle{plainnat} 
\bibliography{references} 

\end{document}